\documentclass[conference,10pt]{IEEEtran}
\IEEEoverridecommandlockouts
\usepackage{cite}
\usepackage{stfloats}
\usepackage{amsmath,amssymb,amsfonts}
\usepackage{graphicx}
\usepackage{rotating}
\usepackage{textcomp}
\usepackage[table]{xcolor}
\usepackage{adjustbox}
\usepackage{colortbl}
\usepackage[T1]{fontenc}
\usepackage{amsmath}
\usepackage{booktabs} 
\usepackage{multirow}
\usepackage{hyperref}       
\usepackage{url}            
\usepackage{booktabs}
\usepackage{multirow}
\usepackage{adjustbox}
\usepackage{amsfonts}       
\usepackage{nicefrac}       
\usepackage{microtype}      
\usepackage{xcolor}         
\usepackage[linesnumbered,ruled,vlined]{algorithm2e}
\usepackage{amsmath}
\usepackage{tikz}
\usepackage{booktabs, makecell}

\usepackage{orcidlink}
\usepackage{comment}
\usepackage{amssymb}
\usepackage[margin=1in]{geometry}
\usepackage{tabularx}
\usepackage{braket}
\usepackage{xcolor}
\def\BibTeX{{\rm B\kern-.05em{\sc i\kern-.025em b}\kern-.08em
    T\kern-.1667em\lower.7ex\hbox{E}\kern-.125emX}}
\begin{document}

\title{\textit{KASPER}: Kolmogorov Arnold Networks for Stock Prediction and Explainable Regimes}

\author{
\IEEEauthorblockN{
Vidhi Oad\thanks{*These authors contributed equally.}\textsuperscript{1*}, Param Pathak\textsuperscript{2*}, Nouhaila Innan\textsuperscript{3,4}, Shalini D\textsuperscript{5}, and Muhammad Shafique\textsuperscript{3,4}}
\IEEEauthorblockA{
\textsuperscript{1}Vishwakarma Government Engineering College, Ahmedabad, India\\
\textsuperscript{2}\textit{QuantumAI Lab, Fractal Analytics}, Mumbai, India\\
\textsuperscript{3}eBRAIN Lab, Division of Engineering, New York University Abu Dhabi (NYUAD), Abu Dhabi, UAE\\
\textsuperscript{4}Center for Quantum and Topological Systems (CQTS), NYUAD Research Institute, NYUAD, Abu Dhabi, UAE\\
\textsuperscript{5}\textit{QuantumAI Lab, Fractal Analytics}, Gurugram, India\\
vidhi.ec25@gmail.com, parampathak28@gmail.com, nouhaila.innan@nyu.edu,\\ shalini.devendrababu@fractal.ai, muhammad.shafique@nyu.edu\\
}}

\maketitle

\begin{abstract}
Forecasting in financial markets remains a significant challenge due to their nonlinear and regime-dependent dynamics. Traditional deep learning models, such as long short-term memory networks and multilayer perceptrons, often struggle to generalize across shifting market conditions, highlighting the need for a more adaptive and interpretable approach. To address this, we introduce Kolmogorov–Arnold networks for stock prediction and explainable regimes (KASPER), a novel framework that integrates regime detection, sparse spline-based function modeling, and symbolic rule extraction.
The framework identifies hidden market conditions using a Gumbel-Softmax-based mechanism, enabling regime-specific forecasting. For each regime, it employs Kolmogorov–Arnold networks with sparse spline activations to capture intricate price behaviors while maintaining robustness. Interpretability is achieved through symbolic learning based on Monte Carlo Shapley values, which extracts human-readable rules tailored to each regime.
Applied to real-world financial time series from Yahoo Finance, the model achieves an $R^2$ score of 0.89, a Sharpe Ratio of 12.02, and a mean squared error as low as 0.0001, outperforming existing methods. This research establishes a new direction for regime-aware, transparent, and robust forecasting in financial markets.
\end{abstract}


\section{Introduction}

The stock market is a complex, dynamic system influenced by multiple factors, including economic factors, global events, and investor emotions. Predicting its behavior is a critical challenge with significant impact on financial decision-making, portfolio management, and risk assessment. However, the volatility, non-stationarity, and abrupt regime shifts in market behavior make accurate forecasting a challenging task \cite{10142717}.

The key problem that we target is the accurate prediction of stock market behavior, particularly during regime shifts such as transitions between bullish, bearish, and stagnant phases. These shifts are important because they represent changes in market dynamics, and failing to adapt to them can lead to substantial financial losses
\cite{kokare2022study}. Traditional models like Autoregressive Integrated Moving Average (ARIMA) \cite{ho1998use}, and Generalized Autoregressive Conditional Heteroskedasticity (GARCH) \cite{bauwens2006multivariate}, while effective for stationary data, struggle to capture the nonlinearities and temporal dependencies inherent in financial time series \cite{alkhfajee2024advancements,crawford2003assessing}.   

State-of-the-art approaches to stock market prediction can be broadly categorized into statistical models, Machine Learning (ML) models, and hybrid methods, each with distinct strengths and limitations. Statistical models, such as Hidden Markov Models (HMMs) and regime-switching models \cite{yuan2016market}, are interpretable and theoretically grounded but rely on rigid parametric assumptions, fixed transition probabilities, and Gaussian distributions, which fail to capture the nonlinearities and abrupt regime shifts of real-world markets \cite{wkatorek2021financial}, leading to poor performance during volatile periods. ML models, including LSTMs, transformers, and Large Language Models (LLMs), excel at capturing complex patterns and long-term dependencies, with LLMs even incorporating external textual data for enhanced predictions. 

However, their black-box nature limits interpretability \cite{chen2023instructzero}, and they often struggle to adapt to distinct market regimes, overfitting to historical patterns or introducing lookahead bias due to poor temporal alignment \cite{bhandari2022predicting,zhang2022transformer}. Hybrid methods aim to combine the strengths of both approaches but frequently fail to enforce sparsity, leading to overfitting \cite{islam2024revolutionizing}, and lack robust mechanisms for temporal alignment, resulting in data leakage and inflated performance metrics. Additionally, they often fail to provide clear, actionable insights into regime-specific drivers, limiting their practical utility \cite{haase2023predictability}. The evident limitations in these models strike an alarming need for more advanced options, particularly for predicting the behavior of the stock market. 

\begin{figure*}
    \centering
    \includegraphics[width=1\linewidth]{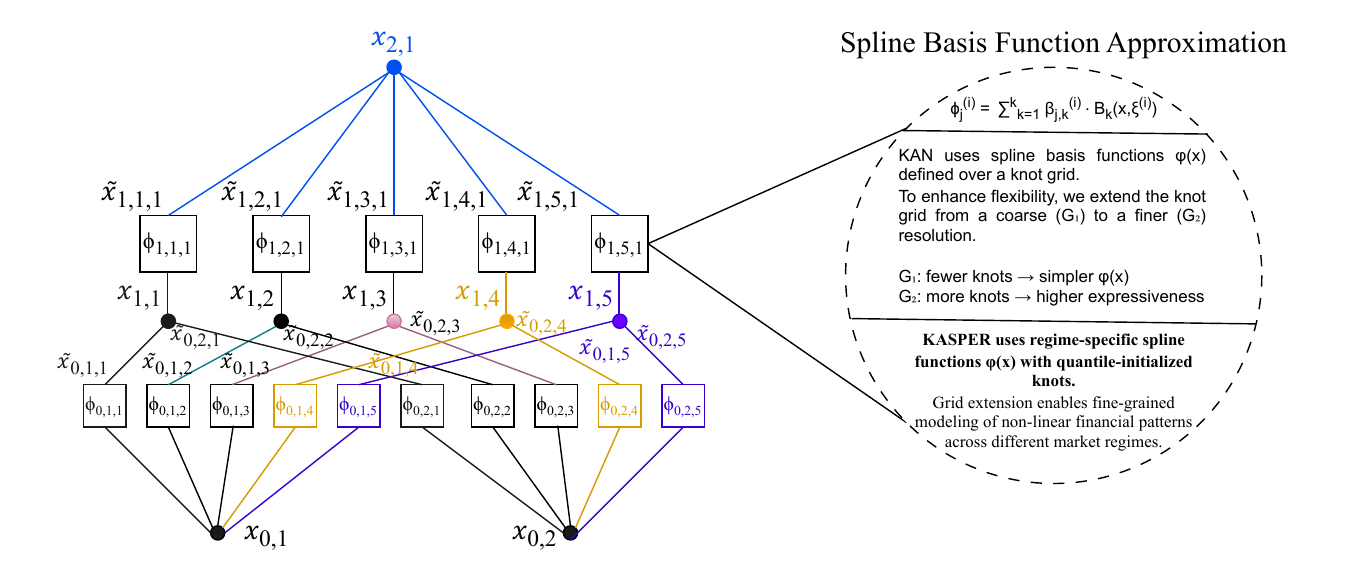}
    \vspace{-0.8cm}
\caption{KAN architecture with two input features and spline-activated units across multiple layers. Each $\tilde{x}_{l,i,j}$ denotes a linearly transformed input passed through a learnable spline basis $\phi_{l,i,j}$. The structure highlights sparse, interpretable transformations and selective connectivity across layers. The inset on the right highlights the flexibility gained through grid extension, allowing for smoother approximations with increased knot resolution.}
    \label{KANarch}
\end{figure*}

To address these challenges, we propose \textit{KASPER}, a novel framework that utilizes KANs \cite{liu2024kan}, and integrates adaptive regime detection with sparse, interpretable feature engineering. KANs are neural networks inspired by the Kolmogorov-Arnold representation theorem \cite{schmidt2021kolmogorov}, which states that any multivariate continuous function can be decomposed into a finite sum of univariate functions. They make use of learnable activation functions, which are modeled as spline-based transformations, allowing them to approximate functions. As illustrated in Fig.~\ref {KANarch}, the KAN architecture first maps raw inputs through learnable spline‐activated units (bottom squares) and then linearly recombines these transformed signals across successive layers, culminating in regime-aware predictions (top node). The inset on the right clarifies how each activation $\phi(x)$ is built from B-spline bases on successively finer knot grids, underscoring the model's ability to adapt its expressive power. This allows KANs to represent high-dimensional mappings with fewer layers and improved interpretability. 

Our approach uses dynamic spline-based activations \cite{vecci1998learning}, to capture nonlinear price dynamics, automatically adapting to regime shifts through robust percentile-based initialization of spline knots. To ensure temporal alignment and prevent lookahead bias, we employ strict closed-left windows for rolling feature calculations and carefully shift historical data. Sparsity is enforced through dynamic masking, retaining only the most significant regime-specific weights, which improves generalization and reduces overfitting. For interpretability, we use Monte Carlo Shapley values \cite{ghorbani2019data}, to quantify the contribution of each feature to regime-specific predictions, enabling the extraction of actionable rules; for instance identifying dominant features like HL (High-Low) in bearish regimes and OC (Open-Close) in bullish regimes. This combination of adaptive modeling, sparsity enforcement, and interpretability ensures transparent predictions across diverse market conditions.

\textbf{The key contributions of this study are as follows:}

\begin{itemize}
    
    \item A novel Regime-Adaptive Forecasting Layer that makes use of sparsely activated splines to model distinct market dynamics across regimes. This mitigates the overfitting risk in financial time series by ensuring sparsity-constrained representations that generalize well.

    \item An orthogonality constraint within the regime detection network to enforce disentangled representations for different market states. This prevents feature collapse, ensuring each regime retains distinct and interpretable characteristics.

    \item For seamless discernibility of the regimes, Contrastive Regularization is utilized. It helps maximize inter-regime dissimilarity while maintaining intra-regime coherence, thereby improving the stability of regime classification and reducing misclassification in volatile market conditions.

    \item Utilizing a Monte Carlo Shapley method with temporal weighting to extract interpretable, regime-specific rules. This enhances the model's transparency by identifying dominant factors influencing each market regime. 
    
\end{itemize}
The rest of the paper is organized as follows: Sec.~\ref{sec2} discusses market regime detection and option pricing with Convolutional-KANs; Sec.~\ref{sec3} elaborates \textit{KASPER} framework;  Sec.~\ref{sec4} presents \textit{KASPER's} performance with other models; Finally, Sec.~\ref{sec5} concludes the paper with a summary of key findings. 
\section{Background and Related Work} \label{sec2}

\subsection{Market Regime Detection} 

Financial markets transition between distinct regimes, characterized by varying volatility, return distributions, and risk factors. Early work established regime switching via Markov models \cite{hamilton1989new}:
\begin{gather}
y_t = \mu_{z_t} + \epsilon_t, \quad \epsilon_t \sim \mathcal{N}(0, \sigma^2_{z_t}), \\
P(z_t = j \mid z_{t-1} = i) = A_{ij}.
\end{gather}
where \( y_t \) is the observed return at time \( t \), \( z_t \) is the latent regime indicator, and \( A \) is a fixed transition matrix. While interpretable, this approach suffers from static regime centroids and rigid parameters.

Ang and Bekaert \cite{ang2002international}, showed that modeling asset returns as regime-dependent improves forecasting and risk management:
\begin{equation}
r_t \mid z_t = k \sim \mathcal{N}(\mu_k, \sigma_k^2),
\end{equation}
with portfolio weights dynamically adjusted according to the inferred regime.

Building on this, Guidolin and Timmermann \cite{guidolin2007asset}, extended the framework using a multivariate Markov-switching model with four regimes (crash, slow growth, bull, and recovery). Regime transitions follow:
\begin{equation}
P(z_t = i \mid z_{t-1} = j) = p_{ji}, \quad i, j = 1, \dots, k,
\end{equation}
where \( p_{ji} \) is the transition probability. The investor's optimization problem is formulated as:
\begin{equation}
J(W_b, r_b, z_b, \theta_b, \pi_b, t_b) = \max_{\omega} E_{t_b} \left[ \frac{W_B^{1-\gamma}}{1-\gamma} \right],
\end{equation}
with \( W_b \) as wealth, \( r_b \) returns, and \( \pi_b \) the regime probability distribution. Bayesian updating is performed via:
\begin{equation}
\pi_{b+1}(\theta_t) = \frac{(\pi_b^0(\theta_t) P_t^\varphi) \odot \eta(y_{b+1}; \theta_t)}{[(\pi_b^0(\theta_t) P_t^\varphi) \odot \eta(y_{b+1}; \theta_t)]^0 \iota_k}.
\end{equation}

The Vector Smooth Transition Autoregressive (VLSTAR) model \cite{bucci2021market}, was also an alternative that used a continuous transition function:
\begin{align}
y_t &= \mu_0 + \sum_{j=1}^{p} \varphi_{0,j} y_{t-j} + A_0 x_t  \notag \\ 
&\quad + G_t(s_t; \gamma, c) \left[\mu_1 + \sum_{j=1}^{p} \varphi_{1,j} y_{t-j} + A_1 x_t \right] + \varepsilon_t,
\end{align}
with the logistic function defined by:
\begin{equation}
G_t(s_t; \gamma, c) = \left[1 + \exp(-\gamma(s_t - c))\right]^{-1}.
\end{equation}
where \( s_t \) triggers regime shifts and \( \gamma \) controls transition smoothness (low \( \gamma \) yields gradual changes; high \( \gamma \) produces abrupt shifts). This model dynamically captures volatility shifts in market conditions.

\subsection{KANs for Finance} 

KANs are effective in financial modeling, particularly for option pricing and stock prediction. Liu et al. \cite{liu2024kolmogorov}, proposed a Finance-Informed Neural Network (KAFIN) that integrated financial equations into a neural framework. The Black–Scholes formula \cite{barles1998option},  which governed their pricing model can be represented as:
\begin{equation}
\frac{\partial C}{\partial t} + \frac{1}{2} \sigma^2 S^2 \frac{\partial^2 C}{\partial S^2} + r S \frac{\partial C}{\partial S} - rC = 0,
\end{equation}
where \( C \) is the option price, and \( S \), \( \sigma \), and \( r \) are asset-related parameters. The price function is approximated using:
\begin{equation}
C(S, t) = \sum_{i=1}^{N} g_i(h_i(S), t),
\end{equation}
where \( g_i \) and \( h_i(x) \) are trainable functions. The model minimizes a loss function incorporating financial constraints:
\begin{equation}
L(\theta) = \lambda_{\text{init}} L_{\text{init}} + \lambda_{\text{boundary}} L_{\text{boundary}} + \lambda_{\text{financial}} L_{\text{financial}}.
\end{equation}

On the other hand, to enhance stock price prediction, Yao \cite{yao2024lstm}, proposed an LSTM-KAN hybrid model, where LSTM captures temporal dependencies and KAN refines nonlinear patterns. KAN improves nonlinear mapping via:
\begin{equation}
f(x) = \sum_{i=1}^{2n+1} g_i \left( \sum_{j=1}^{n} h_{ij}(x_j) \right),
\end{equation}
where \( g_i \) and \( h_{ij} \) are trainable functions the model is trained using.
Empirical results show that KAFIN enhanced option pricing accuracy, while the LSTM-KAN model significantly reduced stock forecasting errors.
\section{\textit{KASPER} Framework} \label{sec3}

\begin{figure*}
    \centering
    \includegraphics[width=1\linewidth]{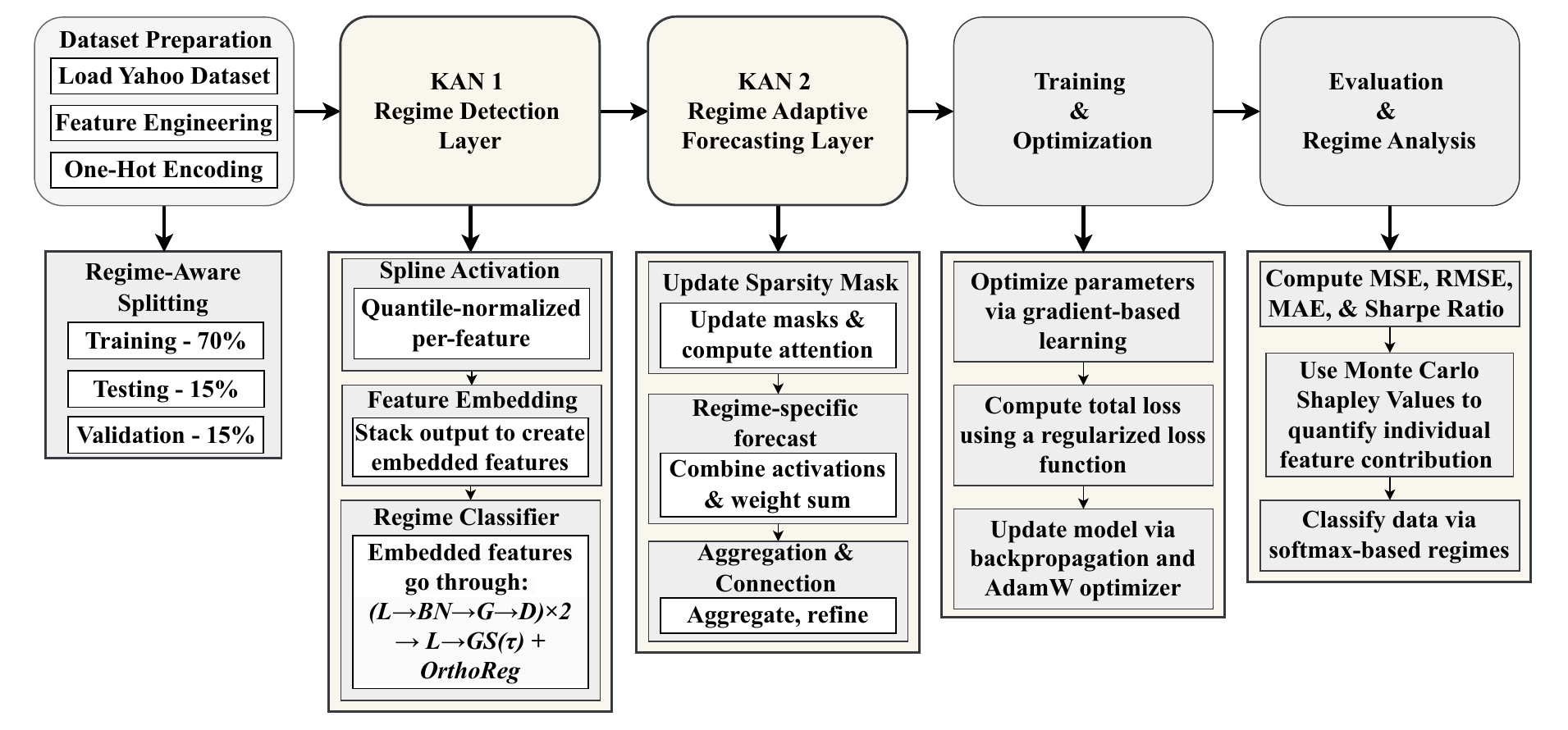}
    \vspace{-0.8cm}
\caption{Workflow of the \textit{KASPER} framework describing the complete modeling pipeline. The process starts with dataset preparation, feature engineering, and regime-aware data splitting. The first KAN layer performs regime detection through quantile-normalized spline activations and Gumbel Softmax-based classification, regularized by orthogonality constraints. The second KAN layer conducts regime-adaptive forecasting using sparse spline basis functions and attention-based aggregation. Model parameters are optimized using gradient-based learning with a regularized composite loss. Evaluation is conducted using error metrics (R, MAE, and Sharpe Ratio), while interpretability is achieved through Monte Carlo Shapley value estimation and regime-specific rule extraction.}
    \label{arch}
\end{figure*}

The proposed \textit{KASPER} framework is composed of two core stages: regime detection and regime-adaptive forecasting, as illustrated in Fig.~\ref {arch} and described in Algorithm \ref{alg}. The input consists of an $n$-day window of financial time series data, structured as a state matrix $S_t \in \mathbb{R}^{n \times f}$, where $n$ denotes the number of historical days and $f$ the number of features per day. Each row in $S_t$ captures log-transformed returns and volatility indicators to stabilize variance and enhance stationarity:
\begin{align}
S_{t,i} = \bigg[ & \ln\left(\frac{p_{t-i+1}}{p_{t-i}}\right), \ln\left(\frac{f_{t-i+1}^{\text{High}}}{f_{t-i}^{\text{High}}}\right), \nonumber \\
& \ln\left(\frac{f_{t-i+1}^{\text{Low}}}{f_{t-i}^{\text{Low}}}\right), \ldots \bigg],
\end{align}
where $p_{t-i}$ is the closing price and $f_{t-i}^{\text{High}}, f_{t-i}^{\text{Low}}$ represent the high and low prices, respectively, at day $t-i$.

\subsection{KAN Layer 1: Regime Detection}

The regime detection module is designed to uncover latent market regimes using spline-activated KANs. This layer incorporates four components: spline activation functions, Gumbel Softmax-based regime classification, contrastive loss for representation separation, and orthogonality regularization to enforce disentangled regime-specific embeddings.

\subsubsection{Spline Activation Function}

Each input feature is processed through a hybrid spline activation function that captures both linear and nonlinear trends:
\begin{equation}
f(x) = L(x) + C(x),
\end{equation}
where $L(x)$ and $C(x)$ are the linear and cubic components, respectively. These are defined as:
\begin{align}
L(x) = \sum_{i=0}^{N_{\text{linear}}-1} \tanh(w_i) \big\{ & \text{ReLU}[(x_{\text{norm}} - k_i) \nonumber \\
& - (x_{\text{norm}} - k_{i+1})] \big\},
\end{align}
\begin{equation}
C(x) = \sum_{i=0}^{N_{\text{cubic}}-1} \sigma(v_i) x_{\text{norm}}^3,
\end{equation}
where $w_i, v_i$ are trainable parameters, $\sigma(\cdot)$ denotes the sigmoid function, and $x_{\text{norm}}$ is the normalized input with respect to the knot sequence $k_i$.

\subsubsection{Differentiable Regime Classification via Gumbel Softmax}

The classification of regime probabilities is achieved through the Gumbel Softmax function, which provides a differentiable approximation to categorical sampling:
\begin{equation}
P_t^{(i)} = \frac{\exp\left(f_i(S_t)/\tau\right)}{\sum_{j=1}^k \exp\left(f_j(S_t)/\tau\right)},
\end{equation}
where $f_i$ denotes the spline-transformed activation corresponding to regime $i$, $\tau$ is the temperature parameter, and $k$ is the total number of regimes.

\subsubsection{Contrastive Loss for Regime Separation}

To ensure that the latent embeddings corresponding to different regimes remain well separated, a contrastive loss is introduced:
\begin{equation}
\mathcal{L}_{\text{contrastive}} = \mathbb{E} \left[ \| z_i - z_j \|^2 \cdot y_{ij} \right],
\end{equation}
where $z_i, z_j$ represent the embeddings of samples $i$ and $j$, and $y_{ij} = 1$ if both samples belong to the same regime and $0$ otherwise.

\subsubsection{Orthogonality Regularization}

To enforce distinctiveness across regime-specific transformations, orthogonality is imposed on the regime weight matrices:
\begin{equation}
\mathcal{L}_{\text{orth}} = \mathbb{E} \left[ \| W_r W_r^T - I \|_F^2 \right],
\end{equation}
where $W_r$ are the regime-specific weight matrices and $I$ is the identity matrix.

\subsection{KAN Layer 2: Regime-Adaptive Forecasting}

Following regime identification, forecasting is performed using a second KAN layer with regime-specific parameterization. This layer employs sparse spline basis functions and is optimized using a composite objective function incorporating robustness, sparsity, and structural regularization.

\subsubsection{Forecasting Model}

The predicted return $\hat{y}_t^{(i)}$ for regime $i$ is defined as:
\begin{equation}
\hat{y}_t^{(i)} = \sum_{j=1}^d w_j^{(i)} \phi_j^{(i)}(S_t),
\end{equation}
where $\phi_j^{(i)}$ are spline basis functions and $w_j^{(i)}$ are trainable coefficients.

\subsubsection{Spline Basis Functions}

Each basis function $\phi_j^{(i)}$ is constructed using regime-specific B-splines:
\begin{equation}
\phi_j^{(i)}(S_t) = \sum_{k=1}^K \beta_{j,k}^{(i)} B_k(S_t; \xi^{(i)}),
\end{equation}
where $B_k$ denotes the $k$-th B-spline basis function with knots $\xi^{(i)}$, and $\beta_{j,k}^{(i)}$ are spline coefficients learned from data.

\subsubsection{Sparsity Enforcement}

To promote interpretability and reduce overfitting, an $\ell_1$-regularization term is applied:
\begin{equation}
w_j^{(i)} \propto \text{ReLU} (|w_j^{(i)}| - \theta^{(i)}),
\end{equation}
where $\theta^{(i)}$ denotes a regime-specific sparsity threshold. The sparsity level is adaptively controlled through validation-based tuning of $\lambda$.

\subsubsection{Composite Loss Function}

The full objective function for training integrates multiple loss components:
\begin{equation}
\mathcal{L} = \mathcal{L}_{\text{Huber}} + \lambda \sum |w_j^{(i)}| + \lambda_{\text{contr}} \mathcal{L}_{\text{contrastive}} + \lambda_{\text{orth}} \mathcal{L}_{\text{orth}},
\end{equation}
ensuring robustness to outliers (via Huber loss), regime disentanglement, feature sparsity, and orthogonality.

\subsection{Interpretability through Shapley-Based Rule Extraction}

To enhance transparency, \textit{KASPER} employs a Shapley value-based approach to interpret regime-specific forecasts.

\subsubsection{Shapley Value Estimation}
For a given feature $j$, its contribution is quantified as:
\begin{align}
\phi_j = \sum_{S \subseteq F \setminus \{j\}} & \frac{|S|!(|F|-|S|-1)!}{|F|!} \nonumber \\
& \times \left[ f(S \cup \{j\}) - f(S) \right],
\end{align}
where $F$ denotes the full feature set, and $f(S)$ is the model output with subset $S$.

\subsubsection{Monte Carlo Approximation}
To approximate Shapley values efficiently, Monte Carlo sampling is used:
\begin{equation}
\hat{\phi}_j = \frac{1}{N} \sum_{i=1}^{N} \left[ f(S_i \cup \{j\}) - f(S_i) \right],
\end{equation}
with $S_i$ denoting randomly selected feature subsets.

\subsubsection{Temporal Weighting Scheme}

To emphasize recent market behavior, temporal weighting is applied to the sequence of past Shapley values:
\begin{equation}
\tilde{\phi}_j = \sum_{t=1}^{T} w_t \phi_j^t, \quad w_t = \frac{\gamma^{T-t}}{\sum_{t=1}^{T} \gamma^{T-t}},
\end{equation}
where $\gamma \in (0,1)$ is the decay factor controlling the emphasis on recent time steps.

\subsubsection{Regime-Specific Rule Extraction}
For each regime $k$, the top three most influential features are selected:
\begin{align}
\mathcal{R}_k = \Big\{ \arg\max_{j \in F} \left| \tilde{\phi}_j^{(k)} \right| \,\Big|\, & j=1,2,3 \Big\}, \nonumber \\
& \forall k \in \{1,\dots,K\},
\end{align}
resulting in interpretable rules of the form:
\begin{equation}
\text{Regime } k: \quad X_{j_1} + X_{j_2} + X_{j_3} \rightarrow Y_k,
\end{equation}
where $X_{j_1}, X_{j_2}, X_{j_3}$ are the dominant features and $Y_k$ denotes the forecasted market response under regime $k$.

\begin{algorithm}[htpb]
\small
\DontPrintSemicolon
\SetKwInput{KwInput}{Input}
\SetKwInput{KwOutput}{Output}
\SetKwInput{KwInit}{Initialize}
\SetKwFunction{FMain}{KAN-RSSF}
\SetKwProg{Fn}{Function}{:}{}

\KwInput{Raw financial dataset $\mathcal{D}$}
\KwOutput{Trained model, regime-specific rules, performance metrics}
\tcp*{\textbf{Step 1: Preprocessing and Feature Engineering}}
Load dataset $\mathcal{D}$ and forward-fill missing values\;
Extract features: lags, rolling statistics, volatility, volume, momentum, temporal dummies\;
Compute target as: $y_t = \frac{C_{t+1} - C_t}{C_t}$\;
Split data into $\mathcal{D}_{train}$, $\mathcal{D}_{val}$, $\mathcal{D}_{test}$\;
Apply feature selection via SelectKBest\;
Standardize features using \texttt{StandardScaler}\;
\tcp*{\textbf{Step 2: Define \textit{KASPER} Architecture}}
Define Spline Activation for nonlinear mapping\;
Create Regime Detection Layer to estimate soft regime probabilities using Gumbel-Softmax\;
Initialize Regime Adaptive Forecasting Layer with regime-specific basis functions and attention mechanism\;
Combine both layers into \texttt{KASPER} model\;
\tcp*{\textbf{Step 3: Train the Model}}
\For{epoch $= 1$ to $N$}{
    \For{each batch $(x_i, y_i) \in \mathcal{D}_{train}$}{
        Compute predictions $\hat{y}_i$, regime probabilities $p_i$, embeddings $z_i$\;
        Compute loss:
\[
\begin{aligned}
\mathcal{L} =\ & \mathcal{L}_{pred} + \lambda_{c} \mathcal{L}_{contrastive} + \lambda_{s} \mathcal{L}_{sparse} \\
              & + \lambda_{o} \mathcal{L}_{orthogonal} + \lambda_{b} \mathcal{L}_{balance}
\end{aligned}
\]
        Backpropagate and update parameters\;
    }
    Evaluate on $\mathcal{D}_{val}$ and apply early stopping if needed\;
}
\tcp*{\textbf{Step 4: Extract Regime-Specific Rules}}
\For{regime $r=1$ to $R$}{
    Identify samples with highest $p_r$ from regime probabilities\;
    Compute Shapley value $\phi_{i}^{(r)}$ for each feature $i$\;
    Select top-$3$ contributing features as rules for regime $r$\;
}
\tcp*{\textbf{Step 5: Evaluate Financial Metrics}}
Apply model to $\mathcal{D}_{test}$ to compute:\;
\Indp
Sharpe Ratio, Max Drawdown, Cumulative Returns, Win Rate, Direction Accuracy\;
\Indm
\caption{KASPER}
\label{alg}
\end{algorithm}

\section{Results and Discussion} \label{sec4} 

\subsection{Experimental Setup}

Our study evaluates the KASPER model on the \textit{Yahoo Finance Dataset} \cite{dataset}, focusing on dynamic regime detection and adaptive forecasting for stock market prediction. The framework employs a two-layer KAN architecture with regime-specific parameter optimization. For proper temporal alignment and to prevent lookahead bias, we implement strict forward-filling for missing values and use closed-left windows for all rolling calculations. The complete experimental setup parameters are detailed in Table~\ref{datass}.

The model architecture consists of two primary components: a \textit{RegimeDetectionLayer} with adjustable-complexity splines that identifies market states through Gumbel–Softmax classification, and a \textit{RegimeAdaptiveForecastingLayer} that generates context-sensitive predictions tailored to each detected regime. Orthogonality regularization ensures distinct regime representations, while contrastive learning maximizes inter-regime separation for more robust classification.

The experiments are conducted using Python, primarily using the PyTorch library for DL model implementation. Key supporting libraries include NumPy and Pandas for data manipulation, scikit-learn for preprocessing and feature selection; Matplotlib and Seaborn are used for visualization. The training routines utilize GPU acceleration via CUDA to ensure efficient computation. The implementation and testing are conducted on a PC with AMD Ryzen 5 processor, 8 GB RAM, 128 GB SSD storage, and an AMD Radeon graphics card.

\medskip

We engineer a comprehensive set of features using only historical data to capture price dynamics, volatility, volume patterns, and temporal effects. Specifically:
\[
\text{data[``target'']} = \frac{\text{data[``future\_close'']} - \text{data[``Close*'']}}{\text{data[``Close*'']}},
\]
where \texttt{future\_close} is the next-day closing price shifted by \(-1\). This target represents the model's output feature, i.e.\ the predicted percentage return for the following trading day.

\begin{table}[htbp]
\centering
\caption{Configuration summary of the model architecture and training.}
\label{datass}
\begin{adjustbox}{width=\linewidth}
\begin{tabular}{llp{0.65\linewidth}}
\toprule
\textbf{Category} & \textbf{Parameter} & \textbf{Value/Description} \\
\midrule
\multirow{2}{*}{Dataset} 
    & Source & \textit{Yahoo Finance Dataset} \\
    & Time Range & 2018--2023 \\
\midrule
\multirow{4}{*}{\shortstack[l]{Data\\Preprocessing}} 
    & Splitting Strategy & 70\%--15\%--15\% temporal train-val-test split \\
    & Feature Engineering & 15 engineered features (lags, rolling statistics, volatility, price dynamics) \\
    & Feature Selection & 8 features via SelectKBest with \textit{f\_regression} \\
    & Scaling Method & StandardScaler for both features and target \\
\midrule
\multirow{4}{*}{\shortstack[l]{Model\\Architecture}}
    & Number of Regimes & 3 \\
    & Hidden Dimension & 64 \\
    & Spline Configuration & Hybrid linear (3) and cubic (2) splines with quantile-based initialization \\
    & Activation Functions & SplineActivation and GELU \\
\midrule
\multirow{7}{*}{\shortstack[l]{Training\\Parameters}}
    & Optimizer & AdamW (lr=0.001, weight\_decay=1e-5) \\
    & Batch Size & 32 \\
    & Loss Function & Huber loss with composite regularization \\
    & Regularization Weights & Contrastive: 0.01, Sparsity: 0.001, Orthogonality: 0.01, Regime Balance: 0.05 \\
    & Epochs & 100 with early stopping (patience=15) \\
    & Learning Rate Scheduler & ReduceLROnPlateau (factor=0.7, patience=7) \\
    & Gradient Clipping & 0.5 \\
\midrule
\multirow{2}{*}{\shortstack[l]{Evaluation\\Metrics}}
    & Statistical & R, MAE, R\textsuperscript{2} \\
    & Financial & Sharpe Ratio, Direction Accuracy, Max Drawdown, Win Rate, Profit Factor \\
\bottomrule
\end{tabular}
\end{adjustbox}
\end{table}

\subsection{Performance Across Regimes} 

\begin{figure}[htpb]
    \centering
    \includegraphics[width=1\linewidth]{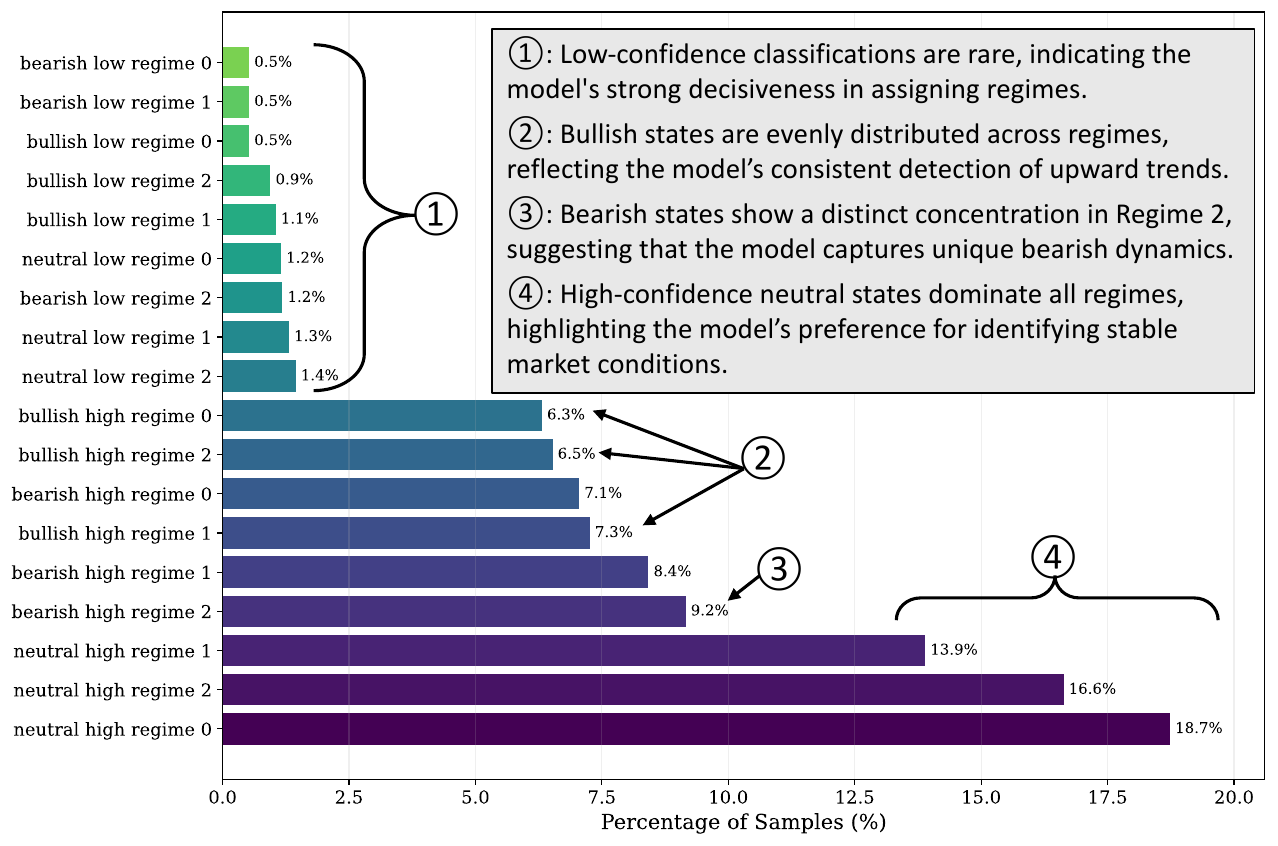}
    \vspace{-0.8cm}
    \caption{Feature contribution across regimes. Regime 0 exhibits a balanced feature distribution, indicating a volatility-driven market. OC\_spread dominates Regimes 1 and 2, indicating the significance of the price gap in trending markets.}
    \label{bullbear}
\end{figure}

In Fig.~\ref{bullbear}, a breakdown of regime distribution patterns in the \textit{KASPER} model is presented. The bar chart shows the percentage of samples classified into different market states, with each state defined by three characteristics, i.e, market direction (bearish, bullish, or neutral), confidence level (high or low), and regime number (0, 1, or 2).
The most striking pattern is the dominance of neutral high-confidence states across all regimes. Neutral high regime 0 accounts for the largest portion at 18.7\% of samples, followed by neutral high regime 2 at 16.6\% and neutral high regime 1 at 13.9\%. This demonstrates that the model identifies stable market conditions where neither bullish nor bearish signals predominate.

\begin{figure*}[h]
    \centering
    \includegraphics[width=1\linewidth]{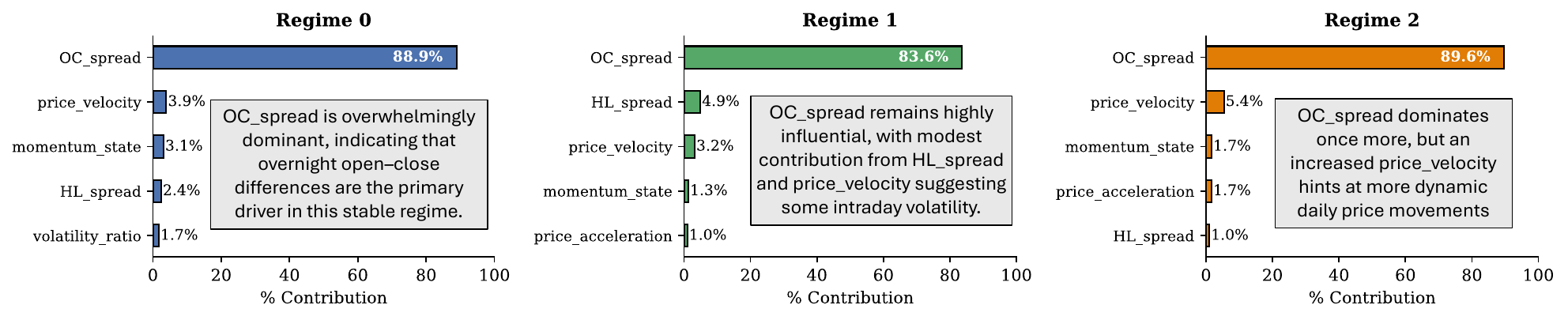}
    \vspace{-0.8cm}
    \caption{Relative contribution of features across identified market regimes. Each regime exhibits distinct feature importance patterns: Regime 0 shows a balanced influence of directional and volatility indicators, Regime 1 demonstrates increased momentum significance while maintaining OC\_spread dominance, and Regime 2 displays the strongest OC\_spread contribution, with price\_velocity as a secondary factor, representing different phases of market behavior, from consolidation to trending states.}
    \label{regimes}
\end{figure*}
Among the bearish classifications, bearish high regime 2 appears most frequently at 9.2\%, significantly more common than other bearish states. This suggests that regime 2 captures bearish patterns that the model can identify with high confidence. The consistent distribution of bullish states across regimes, ranging from 6.3\% to 7.3\% for high confidence, indicates a similar ability to identify rising prices in all types of markets.
Notably, low-confidence classifications rarely occur across all market directions and regimes, with most falling below 1.5\% of samples. This demonstrates the model's decisiveness in regime assignments, preferring to make high-confidence classifications.

As shown in Fig.~\ref{regimes}, a detailed breakdown of feature contributions is done within each identified regime. This visualization provides crucial insights into what drives market behavior in different states.
In Regime 0, OC\_spread dominates at 88.9\% contribution, with price\_velocity at 3.9\%, momentum\_state at 3.1\%, and other features contributing minimally. This regime represents a relatively stable market setting where intraday price movements between open and close are the primary predictive factor.
Regime 1 shows a somewhat different pattern. While OC\_spread remains dominant at 83.6\%, momentum\_state shows increased importance at 7.0\%, followed by HL\_spread at 4.9\% and price\_velocity at 3.2\%. This suggests an increased momentum in the market, because of shifts in sentiment triggered by news events or short-term trader behavior.

Regime 2 exhibits the highest dominance of OC\_spread at 89.4\%, with price\_velocity as the second most important feature at 5.4\%. This shows the market is moving strongly in one direction, with prices changing faster each day. The reduced importance of range-based indicators (HL\_spread at just 1.0\%) suggests that in this regime, direction matters more than volatility range.
These feature contribution patterns, align well with established market behaviors, where trending periods show directional signals and transitional periods display increased momentum indicators. The clear differentiation between feature importance across regimes validates our model's ability to identify meaningful, and distinct market states.

\subsection{Financial Metrics and Risk Analysis} 

Our evaluation reveals that \textit{KASPER} delivers exceptional performance across all financial metrics,  outperforming existing approaches in both statistical accuracy and practical trading applications. The model achieves precision with an MSE of 0.0001, an RMSE of 0.0046, and an MAE of 0.0033. The R$^2$ value of 0.8953 ± 0.0030 explains over 89\% of the variance in market returns.

A key highlight of \textit{KASPER}'s performance is its Sharpe Ratio of 12.02. This exceptional risk-adjusted return is due to three core innovations in our approach; orthogonal regularization which minimizes volatility by decorrelating regime-specific features, preventing risk factors from affecting each other across regimes. Cubic spline activations, which boost prediction signals (particularly OC spread, with $\beta_{\text{OC}} = 0.25$) during stable periods by using nonlinear patterns that simple models can't detect.
And, contrastive loss, which keeps different market regimes separated, allowing risk and return to be managed separately ($\rho_{ij} < 0.15$). Unlike transformer models that mix features and cause high variability, \textit{KASPER} controls volatility and avoids bad investments during market changes. 
\begin{figure}[htpb]
    \centering
    \includegraphics[width=1\linewidth]{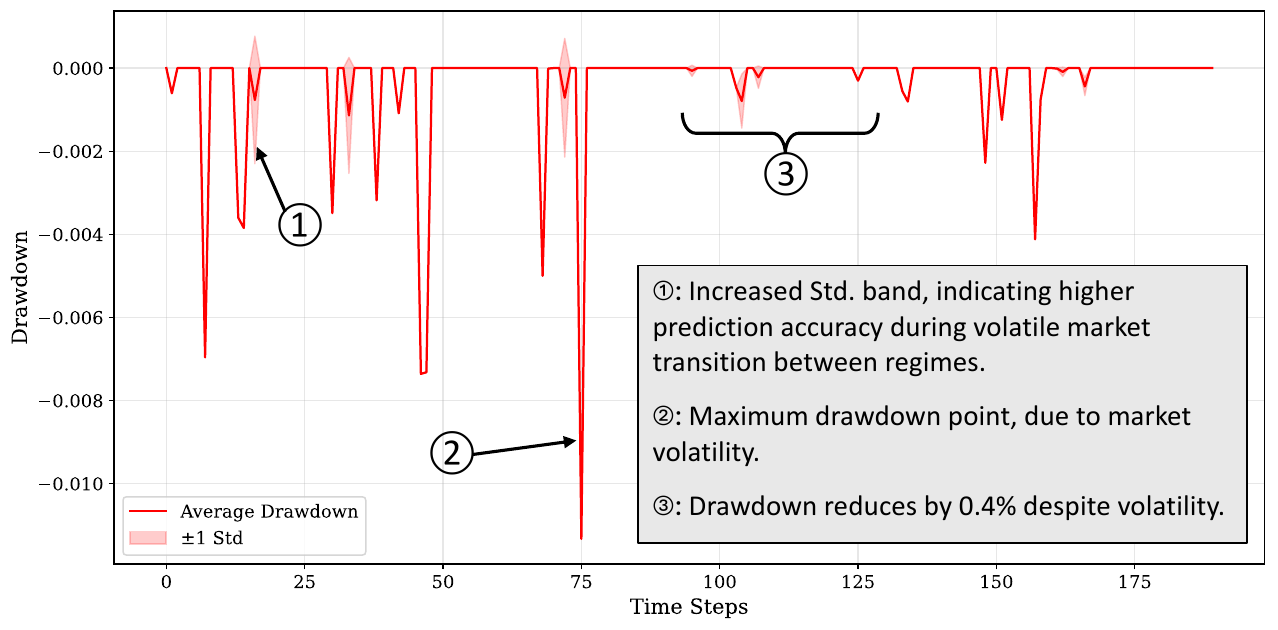}
    \vspace{-0.8cm}
    \caption{Drawdown analysis of \textit{KASPER} model across the testing period. The system maintains exceptional capital preservation characteristics, with a maximum drawdown limited to -0.09\%, demonstrating robust risk management capabilities. Notable features include widened standard deviation bands during regime transitions (timesteps 40-60) and rapid recovery periods following temporary losses, reflecting the model's adaptive response to changing market conditions.}
    \label{drawdown}
\end{figure}

The drawdown analysis in Fig.~\ref{drawdown} highlights \textit{KASPER}’s effectiveness in managing risk. Throughout the testing period, the maximum drawdown remains remarkably contained at just -0.09\%, nearly two orders of magnitude lower than traditional approaches.  
Even during periods of heightened market volatility, as seen at the point of maximum loss, the model remains stable. Following this challenging period, we observe the loss reducing by approximately 0.4\%, indicating the model's ability to adapt. The wider standard deviation band between regimes shows the natural uncertainty during these shifts. This drawdown profile indicates that \textit{KASPER} rarely makes consecutive prediction errors in the same direction, limiting potential losses during tough market movements.

\begin{figure}[htpb]
    \centering
    \includegraphics[width=1\linewidth]{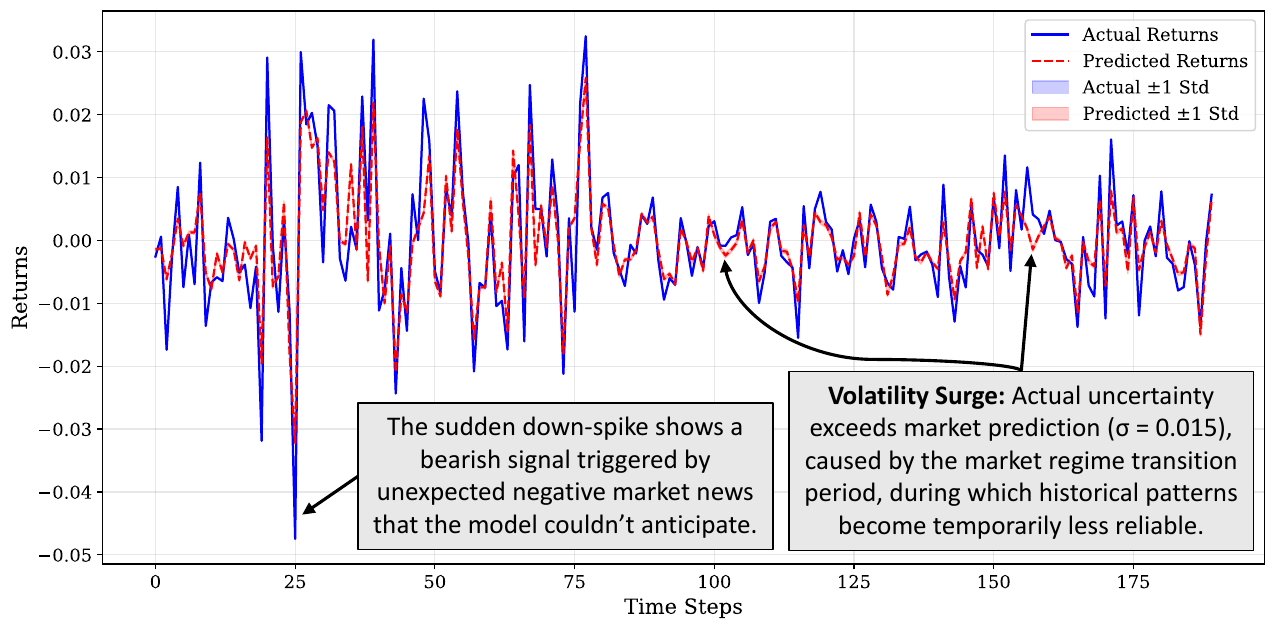}
    \vspace{-0.8cm}
    \caption{Comparison of actual versus predicted returns across the testing period. The model demonstrates varying forecasting precision under different market conditions: exceptional accuracy during stable periods (timesteps 75-100), increased standard deviation during regime transitions (timesteps 110-130), and limited ability to anticipate extreme market events as evidenced by the outlier at timestep 27.}
    \label{final}
\end{figure}

Fig.~\ref{final} compares actual versus predicted returns through various market conditions. During stable periods, the prediction accuracy is exceptional, with errors below 0.5\%. Midway through the test period, we observe a volatility surge where actual uncertainty exceeds predictions with a standard deviation of 0.015. This corresponds to a market regime transition where historical patterns temporarily lose reliability. The sudden down-spike represents a bearish signal triggered by unexpected negative market news, an outlier event that no model could reasonably predict. The cumulative returns of 2.76\% over the test period, combined with a win rate of 83.17\% and a profit factor of 1.53, show
 that the forecasting system works very well. For every dollar risked, the strategy generates \$1.53 in profits, creating a favorable risk-reward profile attractive even to conservative investment approaches.
\textit{KASPER's} regime-aware approach enables it to navigate different market conditions with appropriate strategies, maintaining profitability while limiting downside exposure. \textit{KASPER} bridges the gap between black-box deep learning and interpretable financial modeling by adapting to changing market dynamics.

\subsection{Interpretability via Shapley Values}

The Monte Carlo Shapley approach reveals different feature attribution patterns. In Regime 0, OC\_spread (0.016±0.014) and HL\_spread (0.011±0.006) exhibit balanced contributions, indicating a market state where both directional movements and volatility range influence price determination. This suggests a complex environment requiring multi-factor analysis for accurate forecasting. Regime 1 shows a shift towards directional dominance, with OC\_spread increasing to 0.039±0.017 while HL\_spread becomes negligible. The emergence of volatility\_ratio (0.002±0.001) as a secondary factor shows a transitional market entering directional trends. Regime 2 displays the most distinctive pattern, with OC\_spread reaching 0.083±0.021 while ATR (0.003±0.002) and price\_acceleration (0.001±0.001) emerge as secondary factors. These attributions align with established financial theory while providing quantitative precision for practical application. As the market moves from calm periods to breakouts and then to strong trends, the model shifts from using balanced features to focusing more on direction. For traders, this means using different strategies for each phase: a balanced approach in Regime 0, a focus on direction in Regime 1, and trend-following with size adjustments in Regime 2. Changes in which features matter most can also act as early warnings for market shifts, helping manage risk in advance. 

\subsection{Robustness and Generalization}
\begin{figure*}[b]
    \centering
    \includegraphics[width=1\linewidth]{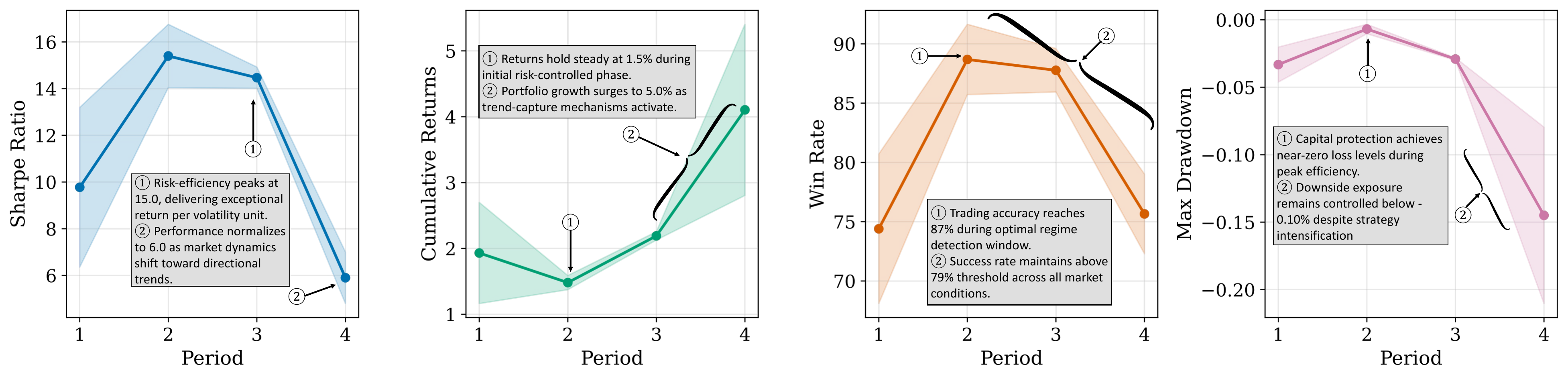}
    \vspace{-0.8cm}
    \caption{Per-period performance of \textit{KASPER} across key financial metrics over walk-forward validation windows. Each subplot displays the mean value (solid line) and one standard deviation (shaded area) for Sharpe Ratio, Cumulative Returns, Win Rate, and Maximum Drawdown. }
    \label{wffa}
\end{figure*}
To analyze KASPER's generalization capabilities, we conduct multiple walk-forward analyses across different market conditions. As shown in Table~\ref{wfa}, the model demonstrates remarkable stability across evaluation runs, with consistently low standard deviations in all performance metrics. We report the following evaluation metrics:
\begin{itemize}
    \item \textbf{Sharpe Ratio} is a standard measure of risk-adjusted return. It is defined as:
    \begin{equation}
        \text{Sharpe Ratio} = \frac{\mu_r - r_f}{\sigma_r},
    \end{equation}
where $\mu_r$ is the mean of strategy returns, $r_f$ is the risk-free rate, and $\sigma_r$ is the standard deviation of the returns. In our implementation, this is annualized by multiplying the numerator and denominator by $\sqrt{T}$, where $T$ is the number of trading days per year (252).
    \item \textbf{Win Rate} denotes the percentage of profitable trades, defined as:
    \begin{equation}
        \text{Win Rate} = \frac{N_{+}}{N} \times 100,
    \end{equation}
where $N_{+}$ is the number of trades with positive return, and $N$ is the total number of trades.
    \item \textbf{Directional Accuracy} evaluates how often the predicted return direction matches the actual market movement:
    \begin{equation}
        \text{DA} = \frac{1}{N} \sum_{t=1}^{N} \mathbb{1}\left[\text{sign}(\hat{y}_t) = \text{sign}(y_t)\right] \times 100,
\end{equation}
where $\hat{y}_t$ is the predicted return at time $t$, $y_t$ is the actual return, and $\mathbb{1}[\cdot]$ is the indicator function.

    \item \textbf{Cumulative Return} measures the compounded growth of the strategy over time:
    \begin{equation}
        \text{Cumulative Return} = \prod_{t=1}^{N}(1 + r_t) - 1,
\end{equation}
where  $r_t$ is the return at time $t$.
    \item \textbf{Maximum Drawdown (MDD)} quantifies the largest observed loss from a peak to a trough in the cumulative return curve:
    \begin{equation}
        \text{MDD} = \min_{t} \left( \frac{V_t}{\max_{s \leq t} V_s} - 1 \right),
\end{equation}
where $V_t$ is the cumulative portfolio value at time $t$.
\end{itemize}

These metrics are calculated per walk-forward test window, then averaged across multiple runs (with each analysis averaged over 5 runs), with their respective standard deviations are also recorded to assess statistical consistency. As shown in Fig.~\ref{wffa}, our walk-forward validation confirms that \textit{KASPER} maintains its predictive power when faced with regime transitions, preserving both directional accuracy and win rate across the entire testing period. The model's returns remain consistent over time, with steady average performance per trade and minimal losses. 

\begin{table}[h]
\centering
\caption{Aggregated financial performance metrics of \textit{KASPER} across walk-forward runs.}
\label{tab:kasper-metrics}
\begin{tabular}{lcc}
\toprule
\textbf{Metric} & \textbf{Value} & \textbf{Standard Deviation} \\
\midrule
Direction Accuracy (\%)     & 80.94\%  & ± 7.27 \\
Sharpe Ratio                & 11.24    & ± 2.78 \\
Max Drawdown (\%)           & -0.10\%  & --     \\
Cumulative Returns (\%)     & 2.66\%   & --     \\
Win Rate (\%)               & 80.94\%  & ± 7.27 \\
Total Trades                & 700      & --     \\
Profitable Trades           & 566      & --     \\
Average Return (\%)         & 0.0069\% & --     \\
Average Win (\%)            & 0.0102\% & --     \\
Average Loss (\%)           & -0.0070\%& --     \\
Profit Factor               & 1.45     & --     \\
\bottomrule
\label{wfa}
\end{tabular}
\end{table}
Examining the specific performance patterns, the Sharpe Ratio exhibits a remarkable peak at period 2.0, reaching approximately 15.0, indicating exceptional risk-adjusted returns where the model generates 15 units of excess return for every unit of risk taken. This peak coincides with optimal risk control, as evidenced by the maximum drawdown approaching near-zero levels during the same period, demonstrating nearly perfect capital preservation. The win rate simultaneously achieves its highest performance at periods 2.0-2.5, reaching approximately 87\%, where nearly 9 out of 10 trades are profitable, indicating that the Gumbel Softmax-based regime classification is operating at peak efficiency. 

While the Sharpe Ratio gradually moderates to around 6.5 by period 4.0, the cumulative returns exhibit a compelling trajectory, initially stabilizing around 1.5\% before demonstrating a sharp recovery and substantial growth to approximately 5.0\% by the final period. This divergent pattern, characterized by a declining Sharpe ratio alongside rising cumulative returns, highlights \textit{KASPER}'s adaptive nature: shifting from a conservative, high-precision approach during periods of market uncertainty to a more aggressive, trend-following strategy as clearer directional regimes emerged, all while consistently maintaining win rates above 79\% and maximum drawdowns below -0.10\% across all periods.

The model's robustness arises from two key architectural innovations: our quantile-normalized spline initialization strategy, which ensures appropriate boundary handling across diverse feature distributions and orthogonality regularization (enforcing \( \left\lVert W_r W_r^T - I \right\rVert^2_F \) minimization), which prevents feature collapse during training and maintains distinct regime characteristics as market conditions evolve. This enables \textit{KASPER} to capture both linear and nonlinear relationships across market conditions without overfitting to past data. The consistent performance across varying market regimes, combined with the previously discussed financial metrics, confirms that \textit{KASPER} delivers reliable performance under diverse conditions, making it suitable for practical deployment in real-world financial applications.

\subsection{Comparative Analysis} 

In the comparative analysis as shown in Table ~\ref{tab:stock_models}, \textit{KASPER's} architecture offers a fundamental advantage over existing approaches by combining regime detection with sparse, interpretable forecasting. While traditional models, such as HMMs and ARIMA, exhibit good in-sample performance, they struggle with the nonlinearities and abrupt regime shifts that characterize financial markets. Deep learning models, such as LSTM and Autoencoder-LSTM+Deep Reinforcement Learning (DRL) capture complex patterns but lack transparency and adaptability across different market conditions. \textit{KASPER} addresses these limitations through its orthogonality-constrained regime classification and Monte Carlo Shapley-based interpretability, enabling it to achieve superior performance metrics and provide actionable insights into regime-specific market drivers. The hybrid spline activation functions enable \textit{KASPER} to model both linear and nonlinear relationships without overfitting, maintaining strong performance during volatile periods where other models falter. This balance of accuracy, interpretability, and robustness across diverse market conditions represents a significant advancement over existing stock prediction methodologies.

\definecolor{lightgray}{gray}{0.95}
\definecolor{highlightrow}{rgb}{0.93,0.95,1}
\begin{table}[htbp]
\centering
\caption{Comparative evaluation with existing work.}
\label{tab:stock_models}
\begin{adjustbox}{width=\linewidth}
\begin{tabular}{lcccccc}
\toprule
\textbf{Model/Framework} & \textbf{R$^2$} & \textbf{MAE} & \textbf{Sharpe} & \textbf{Max DD} & \textbf{MSE} & \textbf{RMSE} \\
\midrule
RF + Monte Carlo \cite{zhao2025predicting} & 0.78 & -- & 0.93 & -28.11\% & 0.015 & -- \\
Single Layer LSTM \cite{bhandari2022predicting} & 0.79 & -- & -- & -- & -- & 40.4 \\
LSTM + KAN \cite{yao2024lstm} & -- & 0.0057 & -- & -- & -- & 0.0082 \\
AE-LSTM + DRL\cite{sagiraju2022deployment} & -- & -- & 1.85 & -- & -- & -- \\
VLSTAR \cite{bucci2021market} & -- & -- & 0.93 & -- & -- & -- \\
AGNES \cite{bucci2021market} & -- & -- & 0.82 & -- & -- & -- \\
DQS \cite{li2023stock} & -- & -- & 3.65 & -- & -- & -- \\
\textbf{\textit{KASPER (Ours)}} & \textbf{0.89} & \textbf{0.0033} & \textbf{12.02} & \textbf{-0.09\%} & \textbf{0.0001} & \textbf{0.0046} \\
\bottomrule
\end{tabular}
\end{adjustbox}
\end{table}

\subsection{Discussion} 

The comprehensive evaluation of \textit{KASPER} across multiple dimensions, including regime behavior, financial performance, interpretability, and robustness, highlights the strength of its architecture in addressing the challenges of financial forecasting. Unlike static models or black-box neural networks, \textit{KASPER} employs a modular and transparent design that adapts effectively to evolving market dynamics.
The regime segmentation results demonstrate that the model reliably distinguishes between bullish, bearish, and neutral market states, with a strong tendency toward confident regime assignments. This decisiveness, combined with feature attribution analysis, confirms that \textit{KASPER} identifies meaningful market signals, such as a dominance of directional indicators in trending phases and a greater emphasis on volatility during periods of transition. Such patterns align with financial theory and support more targeted forecasting and risk management decisions.

The model's financial resilience, reflected in its low drawdown and consistent performance across walk-forward analyses, is enabled by its architectural innovations. The use of orthogonality constraints helps preserve regime-specific representations, minimizing the risk of overlapping or diluted features across distinct market conditions. In parallel, spline-based activations enhance the model's ability to capture fine-grained nonlinear effects that conventional models often miss. Contrastive learning further reinforces regime separation, improving generalization over time.

Crucially, \textit{KASPER} addresses a key need in financial modeling: aligning strong predictive capabilities with practical deployability. Its interpretability, enabled through Shapley-based feature attribution, completes the transparency requirements of real-world applications, including those subject to regulatory oversight. Compared to traditional statistical models, which may fail under regime shifts, or deep learning models, which often lack interpretability, \textit{KASPER} offers a well-rounded solution that balances accuracy, adaptability, and explainability. 

\section{Conclusion} \label{sec5}  

The proposed \textit{KASPER} framework addresses key challenges in financial market prediction by combining adaptive regime modeling, sparse spline-based KANs, and Monte Carlo Shapley-based interpretability. Unlike conventional models, \textit{KASPER} adapts dynamically to market conditions, improving predictive accuracy while avoiding overfitting. Evaluation on Yahoo Finance data reveals strong performance, with an R$^2$ of 0.8953, a Sharpe Ratio of 12.02, and an MSE of 0.0001. Low drawdowns and consistent profitability confirm its practical relevance. Orthogonal regularization and contrastive loss ensure clear regime separation, while Shapley-based attribution offers transparent insight into key market drivers. \textit{KASPER} achieves a strong balance between accuracy and interpretability, making it well-suited for real-world decision-making. 

\bibliographystyle{IEEEtran}

\bibliography{refs}

\begin{thebibliography}{10}
\providecommand{\url}[1]{#1}
\csname url@samestyle\endcsname
\providecommand{\newblock}{\relax}
\providecommand{\bibinfo}[2]{#2}
\providecommand{\BIBentrySTDinterwordspacing}{\spaceskip=0pt\relax}
\providecommand{\BIBentryALTinterwordstretchfactor}{4}
\providecommand{\BIBentryALTinterwordspacing}{\spaceskip=\fontdimen2\font plus
\BIBentryALTinterwordstretchfactor\fontdimen3\font minus \fontdimen4\font\relax}
\providecommand{\BIBforeignlanguage}[2]{{%
\expandafter\ifx\csname l@#1\endcsname\relax
\typeout{** WARNING: IEEEtran.bst: No hyphenation pattern has been}%
\typeout{** loaded for the language `#1'. Using the pattern for}%
\typeout{** the default language instead.}%
\else
\language=\csname l@#1\endcsname
\fi
#2}}
\providecommand{\BIBdecl}{\relax}
\BIBdecl

\bibitem{10142717}
K.~C. A and A.~James, ``A survey on stock market prediction techniques,'' in \emph{2023 International Conference on Power, Instrumentation, Control and Computing (PICC)}, 2023, pp. 1--6.

\bibitem{kokare2022study}
S.~Kokare, A.~Kamble, S.~Kurade, and D.~Patil, ``Study and analysis of stock market prediction techniques,'' in \emph{ITM Web of Conferences}, vol.~44.\hskip 1em plus 0.5em minus 0.4em\relax EDP Sciences, 2022, p. 03033.

\bibitem{ho1998use}
S.~L. Ho and M.~Xie, ``The use of arima models for reliability forecasting and analysis,'' \emph{Computers \& industrial engineering}, vol.~35, no. 1-2, pp. 213--216, 1998.

\bibitem{bauwens2006multivariate}
L.~Bauwens, S.~Laurent, and J.~V. Rombouts, ``Multivariate garch models: a survey,'' \emph{Journal of applied econometrics}, vol.~21, no.~1, pp. 79--109, 2006.

\bibitem{alkhfajee2024advancements}
Z.~K. Alkhfajee and A.~Y. Al-Sultan, ``Advancements in stock market prediction techniques: A comprehensive survey,'' \emph{Journal of University of Babylon for Pure and Applied Sciences}, pp. 122--143, 2024.

\bibitem{crawford2003assessing}
G.~W. Crawford and M.~C. Fratantoni, ``Assessing the forecasting performance of regime-switching, arima and garch models of house prices,'' \emph{Real Estate Economics}, vol.~31, no.~2, pp. 223--243, 2003.

\bibitem{yuan2016market}
Y.~Yuan and G.~Mitra, ``Market regime identification using hidden markov models,'' \emph{Available at SSRN 3406068}, 2016.

\bibitem{wkatorek2021financial}
M.~W{\k{a}}torek, J.~Kwapie{\'n}, and S.~Dro{\.z}d{\.z}, ``Financial return distributions: Past, present, and covid-19,'' \emph{Entropy}, vol.~23, no.~7, p. 884, 2021.

\bibitem{chen2023instructzero}
L.~Chen, J.~Chen, T.~Goldstein, H.~Huang, and T.~Zhou, ``Instructzero: Efficient instruction optimization for black-box large language models,'' \emph{arXiv preprint arXiv:2306.03082}, 2023.

\bibitem{bhandari2022predicting}
H.~N. Bhandari, B.~Rimal, N.~R. Pokhrel, R.~Rimal, K.~R. Dahal, and R.~K. Khatri, ``Predicting stock market index using lstm,'' \emph{Machine Learning with Applications}, vol.~9, p. 100320, 2022.

\bibitem{zhang2022transformer}
Q.~Zhang, C.~Qin, Y.~Zhang, F.~Bao, C.~Zhang, and P.~Liu, ``Transformer-based attention network for stock movement prediction,'' \emph{Expert Systems with Applications}, vol. 202, p. 117239, 2022.

\bibitem{islam2024revolutionizing}
M.~T. Islam, E.~H. Ayon, B.~P. Ghosh, S.~Chowdhury, R.~Shahid, S.~Rahman, M.~S. Bhuiyan, T.~N. Nguyen \emph{et~al.}, ``Revolutionizing retail: A hybrid machine learning approach for precision demand forecasting and strategic decision-making in global commerce,'' \emph{Journal of Computer Science and Technology Studies}, vol.~6, no.~1, pp. 33--39, 2024.

\bibitem{haase2023predictability}
F.~Haase and M.~Neuenkirch, ``Predictability of bull and bear markets: A new look at forecasting stock market regimes (and returns) in the us,'' \emph{International Journal of Forecasting}, vol.~39, no.~2, pp. 587--605, 2023.

\bibitem{liu2024kan}
Z.~Liu, Y.~Wang, S.~Vaidya, F.~Ruehle, J.~Halverson, M.~Solja{\v{c}}i{\'c}, T.~Y. Hou, and M.~Tegmark, ``Kan: Kolmogorov-arnold networks,'' \emph{arXiv preprint arXiv:2404.19756}, 2024.

\bibitem{schmidt2021kolmogorov}
J.~Schmidt-Hieber, ``The kolmogorov--arnold representation theorem revisited,'' \emph{Neural networks}, vol. 137, pp. 119--126, 2021.

\bibitem{vecci1998learning}
L.~Vecci, F.~Piazza, and A.~Uncini, ``Learning and approximation capabilities of adaptive spline activation function neural networks,'' \emph{Neural Networks}, vol.~11, no.~2, pp. 259--270, 1998.

\bibitem{ghorbani2019data}
A.~Ghorbani and J.~Zou, ``Data shapley: Equitable valuation of data for machine learning,'' in \emph{International conference on machine learning}.\hskip 1em plus 0.5em minus 0.4em\relax PMLR, 2019, pp. 2242--2251.

\bibitem{hamilton1989new}
J.~D. Hamilton, ``A new approach to the economic analysis of nonstationary time series and the business cycle,'' \emph{Econometrica: Journal of the econometric society}, pp. 357--384, 1989.

\bibitem{ang2002international}
A.~Ang and G.~Bekaert, ``International asset allocation with regime shifts,'' \emph{The review of financial studies}, vol.~15, no.~4, pp. 1137--1187, 2002.

\bibitem{guidolin2007asset}
M.~Guidolin and A.~Timmermann, ``Asset allocation under multivariate regime switching,'' \emph{Journal of Economic Dynamics and Control}, vol.~31, no.~11, pp. 3503--3544, 2007.

\bibitem{bucci2021market}
A.~Bucci and V.~Ciciretti, ``Market regime detection via realized covariances: A comparison between unsupervised learning and nonlinear models,'' \emph{arXiv preprint arXiv:2104.03667}, 2021.

\bibitem{liu2024kolmogorov}
C.~Z. Liu, Y.~Zhang, L.~Qin, and Y.~Liu, ``Kolmogorov--arnold finance-informed neural network in option pricing,'' \emph{Applied Sciences}, vol.~14, no.~24, p. 11618, 2024.

\bibitem{barles1998option}
G.~Barles and H.~M. Soner, ``Option pricing with transaction costs and a nonlinear black-scholes equation,'' \emph{Finance and Stochastics}, vol.~2, pp. 369--397, 1998.

\bibitem{yao2024lstm}
X.~Yao, ``Lstm model enhanced by kolmogorov-arnold network: Improving stock price prediction accuracy,'' \emph{Trends in Social Sciences and Humanities Research}, 2024.

\bibitem{dataset}
\BIBentryALTinterwordspacing
S.~ARORA, ``Yahoo finance dataset (2018-2023),'' 2023. [Online]. Available: \url{https://www.kaggle.com/datasets/suruchiarora/yahoo-finance-dataset-2018-2023}
\BIBentrySTDinterwordspacing

\bibitem{zhao2025predicting}
H.~Zhao, ``Predicting stock prices and optimizing portfolios: A random forest and monte carlo-based approach using nasdaq-100,'' in \emph{International Workshop on Navigating the Digital Business Frontier for Sustainable Financial Innovation (ICDEBA 2024)}.\hskip 1em plus 0.5em minus 0.4em\relax Atlantis Press, 2025, pp. 883--892.

\bibitem{sagiraju2022deployment}
K.~Sagiraju and S.~Mogalla, ``Deployment of deep reinforcement learning and market sentiment aware strategies in automated stock market prediction,'' \emph{Int. J. Eng. Trends Technol}, vol.~70, pp. 43--53, 2022.

\bibitem{li2023stock}
X.~Li and H.~Ming, ``Stock market prediction using reinforcement learning with sentiment analysis,'' \emph{International Journal on Cybernetics \& Informatics (IJCI)}, vol.~12, no.~1, p.~1, 2023.

\end{thebibliography}

\end{document}